\documentclass[conference]{IEEEtran}
\IEEEoverridecommandlockouts
\usepackage{cite}
\usepackage{amsmath,amssymb,amsfonts}
\usepackage{algorithm}
\usepackage{algorithmic}
\usepackage{url}
\usepackage{graphicx}
\usepackage{textcomp}
\usepackage{booktabs}
\usepackage{xcolor}
\def\BibTeX{{\rm B\kern-.05em{\sc i\kern-.025em b}\kern-.08em
    T\kern-.1667em\lower.7ex\hbox{E}\kern-.125emX}}
\begin{document}

\title{Large Scale Evaluation of Deep Learning-based Explainable Solar Flare Forecasting Models with Attribution-based Proximity Analysis
}

% \author{\IEEEauthorblockN{Temitope Adeyeha, Chetraj Pandey, Berkay Aydin}
\author{\IEEEauthorblockN{Temitope Adeyeha, Chetraj Pandey, Berkay Aydin}
\IEEEauthorblockA{\textit{Dept. of Computer Science, Georgia State University, Atlanta, US} \\
\{tadeyeha1, cpandey1, baydin2\}@gsu.edu}
\vspace{-20pt}
}

\maketitle

\begin{abstract}
Accurate and reliable predictions of solar flares are essential due to their potentially significant impact on Earth and space-based infrastructure. Although deep learning models have shown notable predictive capabilities in this domain, current evaluations often focus on accuracy while neglecting interpretability and reliability—factors that are especially critical in operational settings. To address this gap, we propose a novel proximity-based framework for analyzing post hoc explanations to assess the interpretability of deep learning models for solar flare prediction. Our study compares two models trained on full-disk line-of-sight (LoS) magnetogram images to predict $\geq$M-class solar flares within a 24-hour window. We employ the Guided Gradient-weighted Class Activation Mapping (Guided Grad-CAM) method to generate attribution maps from these models, which we then analyze to gain insights into their decision-making processes. To support the evaluation of explanations in operational systems, we introduce a proximity-based metric that quantitatively assesses the accuracy and relevance of local explanations when regions of interest are known. Our findings indicate that the models' predictions align with active region characteristics to varying degrees, offering valuable insights into their behavior. This framework enhances the evaluation of model interpretability in solar flare forecasting and supports the development of more transparent and reliable operational systems.

\end{abstract}


\section{Introduction} \label{sec:introduction}
Solar flares are transient events characterized by the rapid release of electromagnetic radiation in the outer layers of the Sun's atmosphere. These phenomena play a central role in space weather forecasting and can significantly impact various stakeholders, including power grids, satellite communications, and aviation. Flares are classified into five categories—A, B, C, M, and X—each further subdivided into a strength rating ranging from 1.0 to 9.9, based on their peak X-ray flux, as outlined by the National Oceanic and Atmospheric Administration (NOAA) \cite{fletcher2011}. M- and X-class flares are particularly important to predict due to their potential to cause substantial disruptions on Earth and in space \cite{yasyukevich2018}.

Active Regions (ARs) on the Sun, distinguished by disturbed magnetic fields, are the primary initiators of solar flares and other space weather events \cite{toriumi2019}. Many solar flare prediction models analyze data across the entire solar disk (full-disk models \cite{pandey2021, pandey2022, pandey2023}), which can make it challenging to pinpoint the relevant ARs responsible for flares. These models, often neural networks, are referred to as "opaque" or "black boxes" due to their complex learning representations, which obscures their prediction rationale. This lack of transparency raises concerns for operational forecasting systems, highlighting the need to ensure that these models are interpretable and their predictions understandable for successful integration into real-world forecasting applications. To address these challenges, empirical methods from broader machine learning research have been adopted to better understand how neural networks make decisions. Post hoc explanations, often referred to as attribution methods (e.g., \cite{selvaraju2017, gbackprop, IntGrad, DeepShap}), aim to interpret model outputs by identifying which input features contributed most to a given prediction. These methods enhance transparency by providing insights into the internal workings of complex models, making them more trustworthy for critical applications. However, challenges remain in quantitatively evaluating the effectiveness and reliability of these attribution methods, which is crucial for ensuring they provide meaningful and actionable explanations in operational flare forecasting.

Recently, several studies \cite{bhattacharjee2020, pandey2023demo, pandey2023, PandeyAIKE2023} have focused on developing explainable models for solar flare prediction. To interpret a convolutional neural network (CNN) trained on AR patches, Bhattacharjee et al. \cite{bhattacharjee2020} employed an occlusion-based method. More recently, \cite{sun2022} evaluated DeepLIFT \cite{Deeplift} and Integrated Gradients \cite{IntGrad} for CNNs trained on central AR patches (within $\pm70^\circ$). Pandey et al. \cite{pandey2023discovery, pandey2023demo} extended this work by employing explainable full-disk flare prediction models and qualitatively analyzing near-limb flare predictions. This was further expanded in \cite{pandey2023} with a questionnaire-based evaluation of explanations for X-class flares, demonstrating the model's effectiveness across both near-limb and central regions and validating its feasibility for operational forecasting. While some studies emphasize model development \cite{pandey2021,pandey2022book,PandeyICMLA2023}, others incorporate explainability techniques \cite{yi2021, sun2022, pandey2023}; however, none offer automated evaluation of explanations.

Building on our prior work \cite{pandey2023, PandeyICMLA2023}, which introduced an explainable full-disk solar flare prediction model using compressed line-of-sight (LoS) magnetograms and evaluated Guided Grad-CAM (GGCAM) \cite{selvaraju2017} explanations with human-in-the-loop evaluations in \cite{pandey2023}, this paper advances the field with a fully automated system for evaluating explanations. Designed to localize predictions to ARs from full-disk models, this system enhances automation, marking a significant step toward improving solar flare forecasting and increasing trust in predictive models.

The primary contributions of this paper are: (i) We introduce a fully automated quantitative approach for analyzing model explanations, (ii) We introduce novel evaluation metrics to define distance-based relationships between model explanations, when the regions of interest are known, and (iii) we present a comprehensive case study comparing the explanations of two  standard CNN-based models from \cite{pandey2023} and \cite{PandeyICMLA2023}, using proximity score and colocation ratio.

% By advancing beyond the limitations of manual, qualitative evaluations, this work provides a robust, scalable solution for integrating explainable AI into operational solar flare forecasting, setting a new standard in the explainable space weather forecasting.

The remainder of this paper is structured as follows: Sec.~\ref{sec:methodology} details our methodology, including data preparation and the proximity-based approach of evaluating explanations. Sec.~\ref{sec:experimental_evaluation} presents our experimental evaluation and results. Sec.~\ref{sec:discussion} discusses the analysis and the implications of our findings for the validity of deep learning models. Finally, Sec.~\ref{sec:conclusion} concludes the paper and outlines directions for future research.

\section{Methodology} \label{sec:methodology}
In this section, we describe our approach for evaluating explanations. We utilized two full-disk deep learning models presented in \cite{pandey2023} and \cite{PandeyICMLA2023} to assess the efficacy of our method. These CNN-based models were trained using full-disk line-of-sight (LoS) magnetograms from 2010 to 2018 to predict $\geq$M-class flares, with a prediction window of 24 hours. Details on the model configurations and their predictive performance can be found in the relevant papers cited above.

\begin{figure}[htbp]
  \centering
   \includegraphics[width=0.99\linewidth]{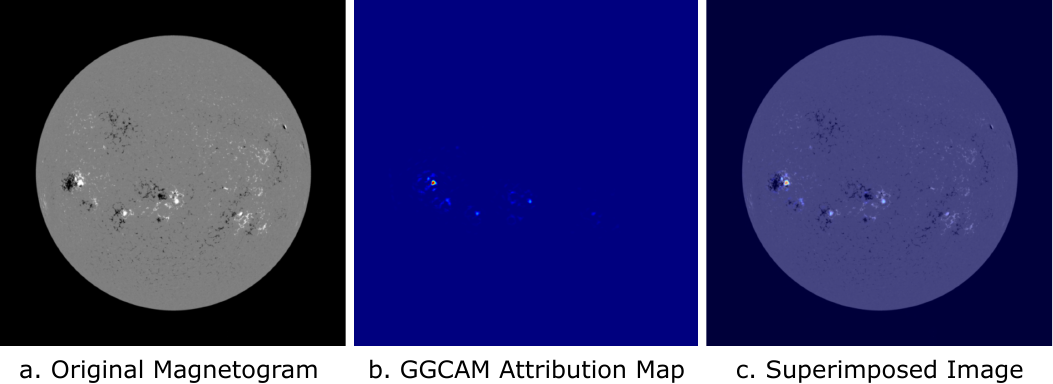}
   \caption{(a) Original input magnetogram image at 2015-01-01T07:59:39.30 UTC. (b) Attribution map generated by Guided Grad-CAM (GGCAM) for input (a). (c) Superimposed image of input (a) and attribution map (b).}
   \label{fig:superimposed}
\end{figure}

For this study, we primarily require full-disk magnetogram images from the Helioseismic and Magnetic Imager (HMI) aboard the Solar Dynamics Observatory (SDO), which represent the Sun's magnetic field strength, with white and black areas indicating positive and negative polarities. These images, as shown in Fig.~\ref{fig:superimposed} (a), are publicly available on Helioviewer\footnote{\url{https://api.helioviewer.org/docs/v2/}}. Furthermore, given an input magnetogram image to the trained full-disk models \cite{pandey2023, PandeyICMLA2023}, we generate attribution maps referred to as explanations. Attribution maps are a type of local post hoc explanation that shows the relative importance of pixels for a given prediction, i.e., the attribution of pixels for a particular decision. We used Guided Grad-CAM (GGCAM) \cite{selvaraju2017}, a gradient-based method that combines Grad-CAM's localization ability with the pixel-level precision of guided backpropagation \cite{gbackprop} to identify important regions in magnetogram images (e.g., see Fig.~\ref{fig:superimposed}(b)). Furthermore, for proximity analysis of explanations corresponding to actual flare locations, we used NOAA-detected flare database\footnote{ftp://ftp.swpc.noaa.gov/pub/warehouse}, which include solar coordinates and associated AR numbers. To ensure a clear comparison between the attribution maps and the magnetograms, preprocessing steps are applied to standardize the attribution maps. Both magnetogram and attribution maps are converted into numpy arrays, normalized, and scaled to 255. For visualization purposes, the arrays are superimposed with an alpha saturation of 0.5, as demonstrated in previous studies \cite{pandey2023} (Fig.~\ref{fig:superimposed} (c)).

\begin{figure*}[h]
  \centering
   \includegraphics[width=0.8\linewidth]{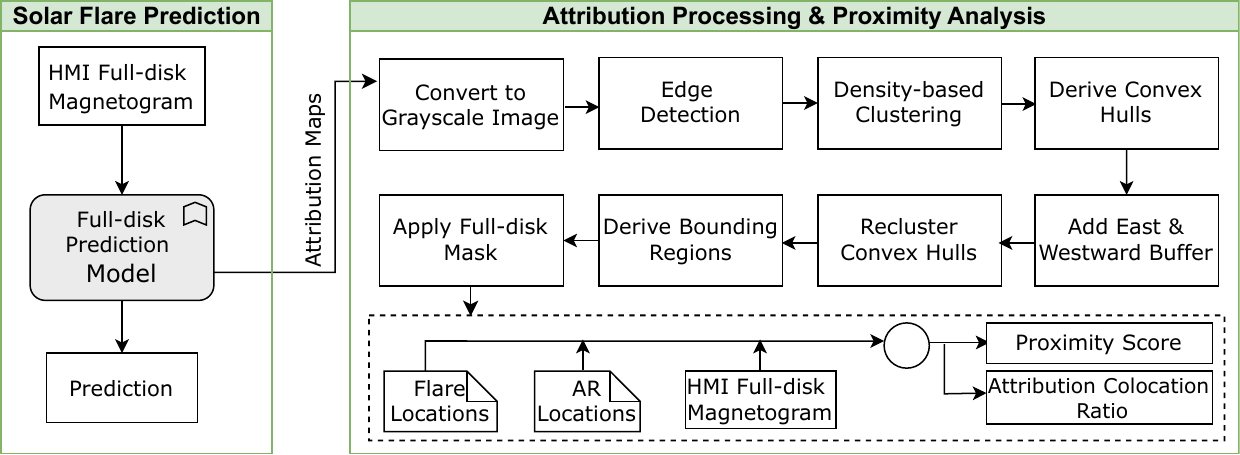}
   \caption{An schematic overview of steps involved in attribution maps preprocessing and corresponding proximity-based analysis used in this study.}
   \label{fig:steps}
   \vspace{-10pt}
\end{figure*}

% \subsection{Explanation Analysis}
\begin{algorithm}[htbp]
\caption{PEPS-SF: Proximity-based Evaluation of Post-hoc Explanations in Solar Flare Prediction}
\label{alg:cap}
\begin{algorithmic}[1]
\renewcommand{\algorithmicrequire}{\textbf{Input:}}
\renewcommand{\algorithmicensure}{\textbf{Output:}}
\REQUIRE $M$ (Magnetogram), $AR$ (AR locations), $FL$ (FL locations)
\ENSURE Proximity Scores ($PS$), Attribution Colocation Ratio ($ACR$)

\textit{Generate Attribution Maps:}
\STATE Attribution Map ($AM$) $\gets \text{GGCAM}(model, M)$

\textit{Scaling (0-255):}
\STATE $G \gets \text{scale}(AM)$

\textit{Edge Detection:}
\STATE $P \gets \text{edge\_detection}(G)$

\textit{Density-Based Clustering:}
\STATE $C, Clusters \gets \text{DBSCAN}(P)$

\textit{Get Convex Hull and Add Buffer Pixels:}
\FOR{each $c \in Clusters$}
    \STATE $CH \gets \text{convex\_hull}(c)$
    \STATE $BR \gets \text{add\_buffer}(CH)$
\ENDFOR

\textit{Coordinate Conversion:}
\STATE $AR_{pix}, FL_{pix} \gets \text{convert\_coordinates}(AR, FL)$

\textit{Proximity Analysis:}
\FOR{each $ar \in AR_{pix}$}
    \STATE $BR_{closest} \gets \text{find\_closest}(ar, BR)$
    \IF{$ar \in BR_{closest}$}
        \STATE $d \gets 0$
    \ELSE
        \STATE $d \gets \text{euclidean\_distance}(ar, BR_{closest})$
    \ENDIF
\ENDFOR

\textit{Calculate Metrics:}
\STATE $PM \gets \text{average}(d)$
\vspace{5pt}
\STATE $ACR \gets \frac{\#AR_{pix} \in BR}{\#AR_{pix}}$
\vspace{5pt}
\RETURN $PS$, $ACR$
\end{algorithmic}
\end{algorithm}
\vspace{-10pt}
\subsection{Preprocessing Attribution Maps}
To quantitatively assess the alignment of highlighted pixels in the attribution maps with observed ARs, we first extract pixel locations and compare them to the AR coordinates. We then apply Canny Edge Detection and density-based clustering to group the important pixels, followed by deriving the convex hull for each group. Next, we compare the coordinates of these convex hulls with the coordinates of the ARs and flares in the original image. To ensure consistency in proximity analysis, we convert spatial coordinates to pixel coordinates, establishing a common scale. This conversion also accounts for the Sun’s rotation, which causes an approximate westward shift of 13 degrees over a 24-hour period, by introducing buffers to the east and west around each convex hull. The entire procedure for proximity-based solar flare prediction and analysis is detailed in Algorithm~\ref{alg:cap} (PEPS-SF), which outlines the generation of attribution maps, scaling, edge detection, density-based clustering, convex hull derivation, and the application of full-disk region (circular mask). Additionally, the algorithm includes the conversion of AR and flare location (FL) coordinates to pixel coordinates, followed by a proximity analysis that calculates the Euclidean distance between each AR and the nearest point on the convex hull. 

The system for solar flare prediction and analysis is composed of interconnected modules as shown in Fig.~\ref{fig:steps}. The Solar Flare Prediction Module initiates the process by processing full-disk magnetogram images, generating prediction and attribution maps that highlight potential solar flare regions, as described in step 1 of Algorithm~\ref{alg:cap}. These maps are then preprocessed by the Attribution Map Processing Module, which executes steps 2 to 9 in Algorithm~\ref{alg:cap}, resulting in a set of maps ready for detailed analysis. For proximity analysis, we use these preprocessed maps and compares them with observational data, including the locations of AR and FL, to calculate the Proximity Score (PS) and Attribution Colocation Ratio (ACR) using steps 10 to 20 in Algorithm~\ref{alg:cap}. This system design provides a structured framework for evaluating how well the activated pixels in the post hoc explanations correspond with actual solar flare activity.
% \vspace{-10pt}

For location-based analysis, heliographic coordinates are converted to pixel coordinates to facilitate accurate proximity analysis, ensuring all spatial data and pixel locations are on a common scale, thereby enabling precise calculations. This conversion is executed using the transformations provided in Eq.~\eqref{eq:pix_x} and Eq.~\eqref{eq:pix_y}:
\vspace{-5pt}
\begin{equation}
\text{pix}_x = \text{HPCCENTER} + \frac{\text{hpcx}}{\text{CDELT}}
\label{eq:pix_x}
\end{equation}
\begin{equation}
\text{pix}_y = \text{HPCCENTER} - \frac{\text{hpcy}}{\text{CDELT}}
\label{eq:pix_y}
\end{equation}

where $\text{pix}_x$ and $\text{pix}_y$ represent the pixel coordinates on the image, $\text{HPCCENTER}$ denotes the center of the solar disk in pixel coordinates, $\text{hpcx}$ and $\text{hpcy}$ are the Helioprojective-Cartesian coordinates, and $\text{CDELT}$ is the scaling factor (i.e., pixel size in degrees). To measure the proximity of active regions to their corresponding convex hulls, the closest convex hull to each active region is identified by computing the Euclidean distance between the region's coordinates and the nearest point on the convex hull, as described by Eq.~\eqref{eq:euclidean_distance}.

\begin{equation}
d_{\text{min}} = \sqrt{(p_x^* - q_x)^2 + (p_y^* - q_y)^2}
\label{eq:euclidean_distance}
\end{equation}

Here, $d_{\text{min}}$ represents the minimum Euclidean distance, with $(q_x, q_y)$ denoting the coordinates of the solar active region and $(p_x^*, p_y^*)$ representing the coordinates of the closest point on the convex hull.

% A working example is provided in Fig.~\ref{fig:process}, which outlines the sequence of transformations applied to an attribution map. The process begins with the \textbf{GuidedGradCAM Image (a)} as the initial attribution map. In \textbf{Stage 1 (b)}, the image is resized and pixel intensities are scaled to 255. This is followed by edge detection in \textbf{Stage 2 (c)} and clustering of the detected edges in \textbf{Stage 3 (d)}. A convex hull is drawn around the clustered points in \textbf{Stage 4 (e)} to define the regions of interest. To account for the Sun's rotation, eastward and westward buffers are added in \textbf{Stage 5 (f)}. In \textbf{Stage 6 (g)}, the locations of the convex hull+buffers are re-clustered to merge smaller clusters introduced by the eastward and westward buffers into larger, more cohesive clusters. A bounding region is then created around these clusters in \textbf{Stage 7 (h)}, and a circular mask is applied in \textbf{Stage 8 (i)} to  isolates the solar disk and removes any extraneous data outside the Sun's circumference. Finally, \textbf{Stage 9 (j)} displays the active region (AR) and flare location (FL) points, indicating whether they fall within the defined bounding regions. This sequence illustrates the proximity-based analysis used to assess the alignment between the model’s predictions and actual solar activity.

\begin{figure*}[h!]
  \centering
   \includegraphics[width=0.75\linewidth]{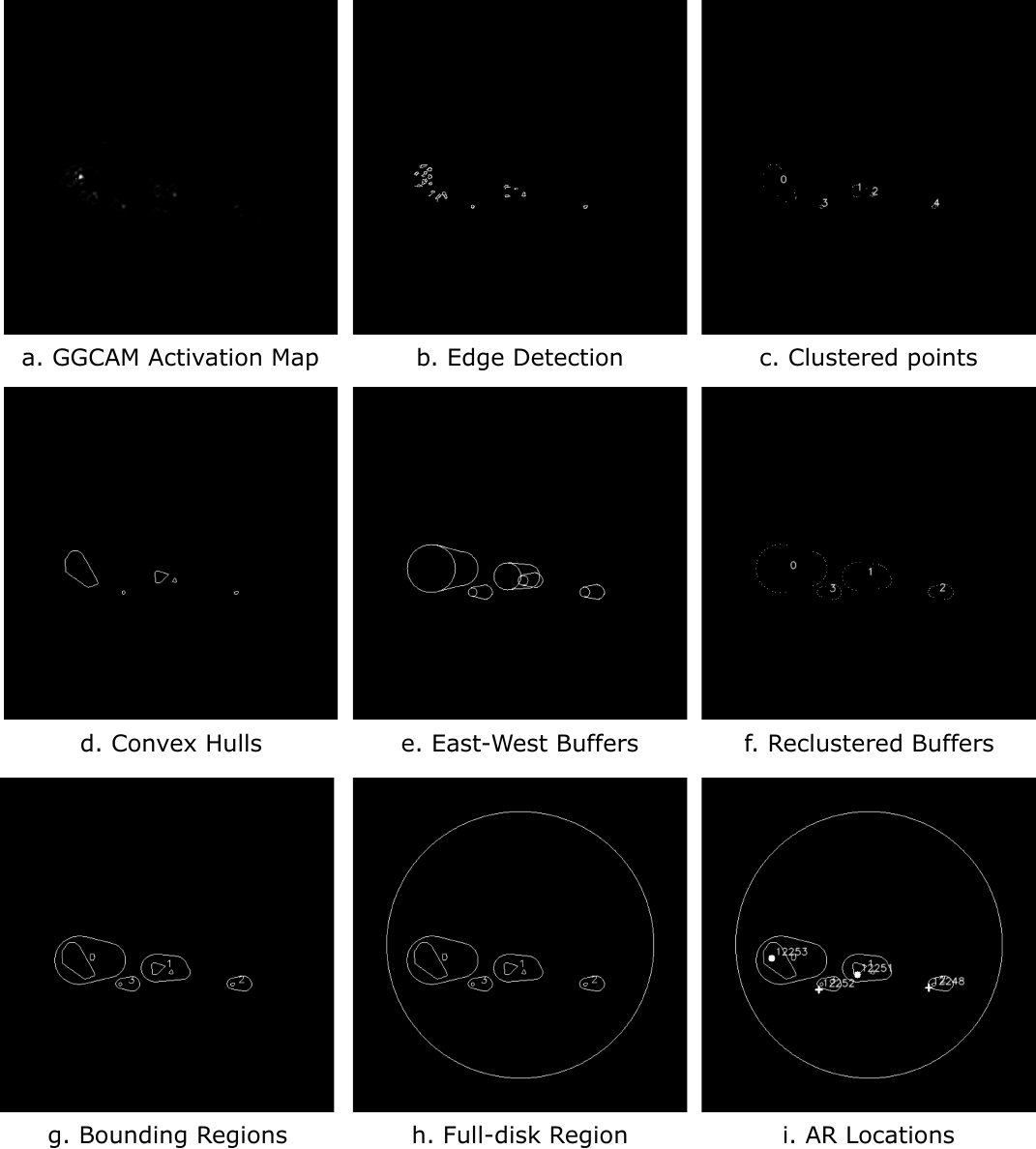}
   \caption{An example of sequential processing applied to the attribution maps of an input magnetogram captured on 2015-01-01 at 07:59:39.30 UTC.}
   \label{fig:process}
   \vspace{-10pt}
\end{figure*}

A working example is provided in Fig.~\ref{fig:process}, which outlines the sequence of transformations applied to an attribution map. The process starts with In \textbf{Step (a)}, the Guided-GradCAM image is resized and pixel intensities are scaled to 255. This is followed by edge detection in \textbf{Step (b)} and clustering of the detected edges in \textbf{Step (c)}. In \textbf{Step (d)}, a convex hull is drawn around the clustered points to define the regions of interest. To account for the Sun's rotation, eastward and westward buffers are added in \textbf{Step (e)}. In \textbf{Step (f)}, these buffered convex hulls are re-clustered to merge smaller clusters into larger, more cohesive clusters. A bounding region is then created around these clusters in \textbf{Step (g)}, and a circular mask is applied in \textbf{Step (h)} to isolate the solar disk, removing any background data outside the Sun's circumference. Finally, \textbf{Step (i)} shows the active region (AR) and flare location (FL) points, indicating whether they fall within the defined bounding regions. This sequence illustrates the proximity-based analysis used to evaluate the alignment between the model’s predictions and actual solar activity.

\subsection{Explanation Evaluation Metrics}
To quantify the explanations' relevance to AR locations, we introduce two metrics: (i) Proximity Score (PS) shown in Eq.~\eqref{eq:proximity_score}, and Attribution Colocation Ratio (ACR) shown in Eq.~\eqref{eq:colocation_ratio}. The PS quantifies the average Euclidean distance between observed ARs and their respective closest convex hulls and ACR measures the percentage of ARs located within the bounding buffer of the total ARs present in full-disk image. To elaborate, PS measures how close the active regions are to the predicted regions, providing insight into the precision of the model's predictions. ACR, on the other hand, indicates whether the active regions are correctly located within the predicted areas. Together, PS offers a continuous measure of distance, while ACR provides a binary measure of correctness. Using both metrics allows for a more comprehensive evaluation, with PS focusing on precision and ACR ensuring that the model captures the relevant areas.

\begin{equation}
PS = \frac{\sum_{i=1}^n d_{\text{min}}(AR_i, AM_i)}{n}
\label{eq:proximity_score}
\end{equation}

\begin{equation}
ACR = \frac{\sum_{i=1}^n \text{inside\_bounding\_box}(AR_i)}{n} \times 100
\label{eq:colocation_ratio}
\end{equation}

where $AR_i$ denotes the $i$th AR, $AM_i$ represents the closest convex hull to $AR_i$,  $n$ is the total number of ARs, and $\text{inside\_bounding\_box}(AR_i)$ is a binary function that equals 1 if $AR_i$ is within the bounding buffer, and 0 otherwise.

\section{Experimental Evaluation} \label{sec:experimental_evaluation}

The focus of this study is to evaluate how well the activated pixels in post hoc explanations align with the locations of active regions (ARs) and flares, particularly X- and M-class flares, which are significant for Earth-impacting events. The proximity-based approach quantifies the quality of these explanations and assesses the models' performance. For our evaluation, the dataset comprises 5,923 full-disk line-of-sight (LoS) solar magnetogram images. These magnetograms, sampled every four hours from January 2020 to December 2023, are labeled with a 24-hour prediction window based on the maximum peak X-ray flux. As mentioned earlier, we used two full-disk models presented in \cite{pandey2023}, denoted as model M1, and model M2, presented in \cite{PandeyICMLA2023}. We used GGCAM-generated attribution maps to perform a proximity analysis, comparing the attribution alignment of the M1 and M2 models with the actual locations of active regions (ARs) and flares. This analysis, guided by the parameters detailed in Table \ref{tab:parameters}, provided critical insights into the models' performance and reliability in predicting solar flares.

\begin{table}[tb]
    \caption{Parameters used for Proximity Analysis of M1 and M2 Models}
    \centering
    \begin{tabular}{@{}lcl@{}}
        \toprule
        \textbf{Parameter} & \textbf{Description} & \textbf{Value} \\
        \midrule
        \text{lower\_threshold} & Lower Pixel Intensity Threshold & 30 \\ 
        \text{upper\_threshold} & Upper Pixel Intensity Threshold & 50 \\
        \text{min\_samples} & Minimum Samples for Clustering & 2 \\
        \text{max\_dist} & Maximum Clustering Distance (pixels) & 10 \\
        \text{eastward\_buffer} & Eastward Buffer (pixels) & 5 \\
        \text{westward\_buffer} & Westward Buffer (pixels) & 40 \\
        \text{target\_size} & Target Image Size (pixels) & 512 \\
        \text{scale\_to} & Pixel Intensity Scale Factor & 255 \\
        \bottomrule
    \end{tabular}
    \label{tab:parameters}
    \vspace{-10pt}
\end{table}

% Table \ref{tab:experimental_other_metrics} summarizes key metrics, including Balanced Accuracy (Bal. Acc.), Precision, Recall, Flare Accuracy (FL Acc.), and Non-Flare Accuracy (NF Acc.) for M1 and M2. While M1 exhibited higher recall and balanced accuracy, M2 showed better precision and non-flare accuracy.

% \begin{table}[tb]
% \centering
% \caption{Other Experimental Metrics for M1 and M2}
% \label{tab:experimental_other_metrics}
% \begin{tabular}{@{}lccccc@{}}
% \toprule
% \textbf{Model} & \textbf{Bal. Acc.} & \textbf{Precision} & \textbf{Recall} & \textbf{FL Acc.} & \textbf{NF Acc.} \\
% \midrule
% M1 & 0.75 & 0.43 & 0.66 & 0.66 & 0.84 \\
% M2 & 0.59 & 0.58 & 0.20 & 0.20 & 0.97 \\
% \bottomrule
% \end{tabular}
% \end{table}

\begin{figure}[tb]
\centering
\includegraphics[width=0.9\linewidth]{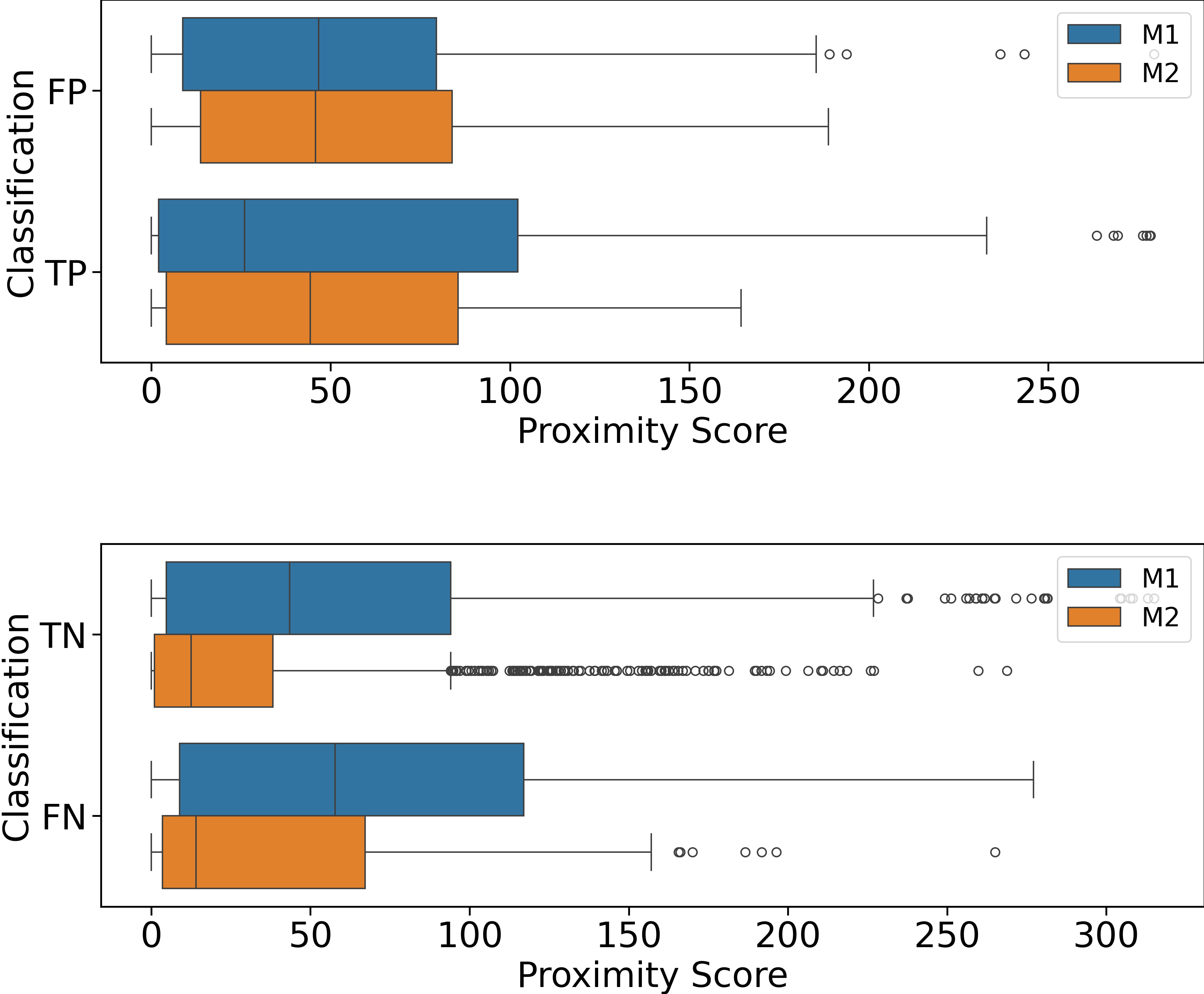}
\caption{Distribution of Proximity Scores, grouped in terms of four outcomes of contingency matrix (referred in text as categories). (a) Proximity score distribution for True Positives (TP) and False Positives (FP) (b) Proximity score distribution for False Negatives (FN) and True Negatives (TN).}
\label{fig:proximity_scores_boxplot}
\end{figure}

While the predictive performance of models M1 \cite{pandey2023} and M2 \cite{PandeyICMLA2023} is reported in the respective studies, we present an evaluation of the explanations in terms of the average Proximity Score (PS) and Attribution Colocation Ratio (ACR) generated using these models, as shown in Table \ref{tab:proximity_colocation_summary}. We observed that M2 demonstrates better spatial alignment with actual flare locations, as indicated by a lower PS, while both models perform similarly in terms of ACR. Fig.~\ref{fig:proximity_scores_boxplot} illustrates the distribution of proximity scores across different classification categories (TN, FP, TP, FN) for M1 and M2 respectively. The results indicate that M2 generally produces lower median proximity scores with less variability, suggesting more consistent and accurate predictions of flare regions compared to M1. The source code and other necessary artifacts are available from our open-source repository \cite{dmlab2024}.

\begin{table}[tb]
\centering
\caption{Average PS and ACR for M1 and M2 models}
\label{tab:proximity_colocation_summary}
\begin{tabular}{@{}lcc@{}}
\toprule
\textbf{Model} & \textbf{Average PS} & \textbf{Average ACR} \\
\midrule
M1 & 61.405091 & 0.552943 \\
M2 & 29.420010 & 0.550404 \\
\bottomrule
\end{tabular}
\vspace{-10pt}
\end{table}

% \begin{figure}[tb]
% \centering
% \includegraphics[scale=0.3]{figure/proximity_score_boxplot_new.png}
% \caption{a) Distribution of Proximity Scores by Classification Category for M1 Model. b) Distribution of Proximity Scores by Classification Category for M2 Model.}
% \label{fig:proximity_scores_boxplot}
% \end{figure}

% \begin{figure*}[tb]
% \centering
% \includegraphics[scale=0.5]{figure/colocation_ratio_boxplot_new.png}
% \caption{a) Distribution of Attribution Colocation Ratios by Classification Category for M1 Model. b) Distribution of Attribution Colocation Ratios by Classification Category for M2 Model.}
% \label{fig:colocation_ratio_boxplot}
% \end{figure*}

\begin{figure}[tb]
\centering
\includegraphics[width=0.9\linewidth]{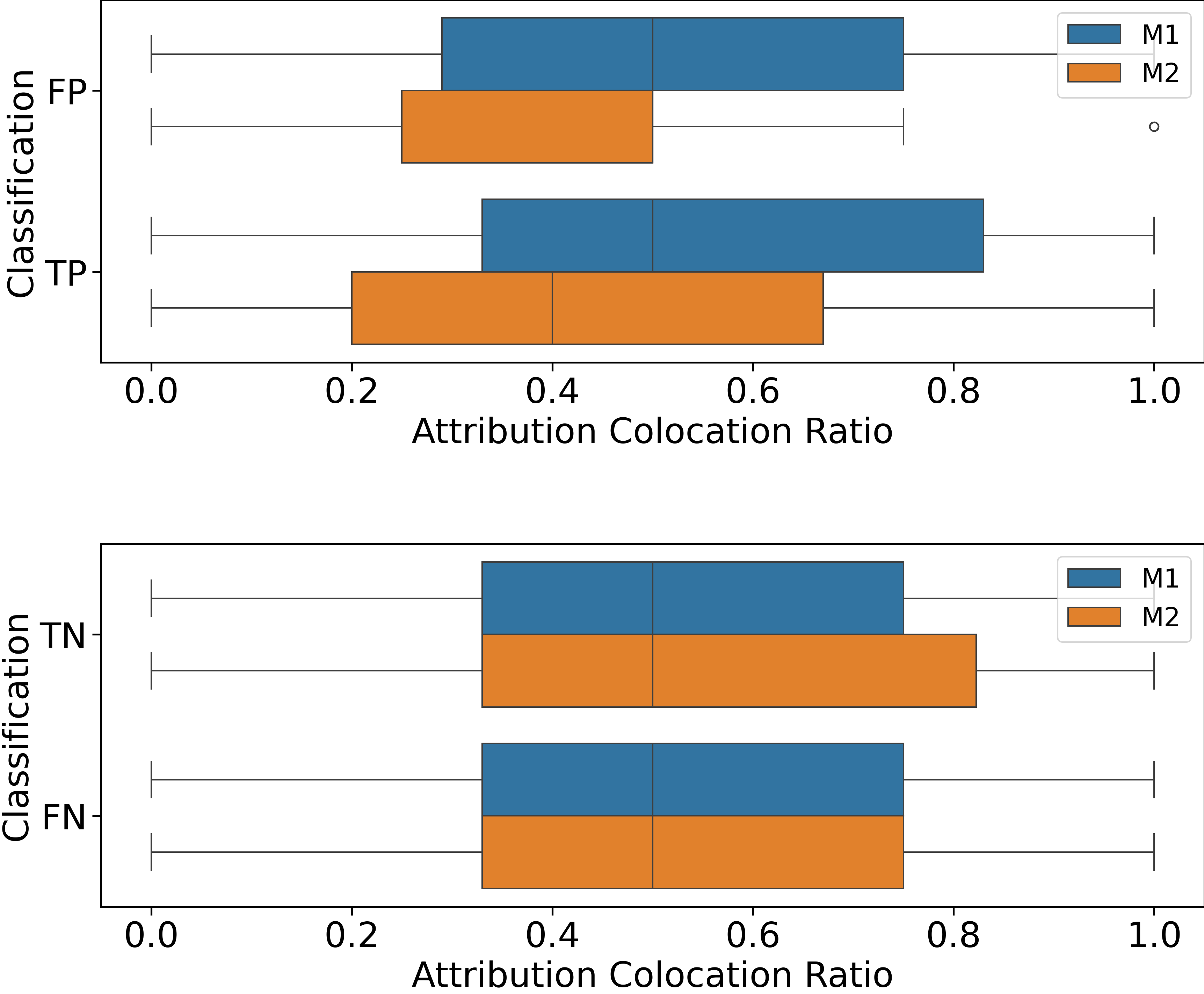}
\caption{Distribution of Attribution Colocation Ratios for both models, grouped in terms of four outcomes of contingency matrix (referred in text as categories). (a) ACR distribution for True Positives (TP) and False Positives (FP) (b) ACR distribution for False Negatives (FN) and True Negatives (TN).}
\label{fig:colocation_ratio_boxplot}
\vspace{-15pt}
\end{figure}

\section{Discussion} \label{sec:discussion}
\subsection{Comparative Analysis of PS Across Models}
\vspace{-5pt}
\begin{table}[htbp!]
\centering
\caption{Summary of Proximity-Based Scores for M1 and M2 Models}
\label{tab:proximity_comparison}
\begin{tabular}{lcccc}
\toprule
\textbf{Metric} & \multicolumn{2}{c}{\textbf{Mean}} & \multicolumn{2}{c}{\textbf{Standard Deviation}} \\ 
 & \textbf{M1} & \textbf{M2} & \textbf{M1} & \textbf{M2} \\ \midrule
\textbf{FN} & 72.49 & 38.02 & 71.89 & 46.96 \\ 
\textbf{FP} & 52.95 & 55.63 & 52.72 & 49.05 \\ 
\textbf{TN} & 61.70 & 28.53 & 66.59 & 40.26 \\ 
\textbf{TP} & 59.43 & 52.03 & 68.05 & 48.68 \\ 
\bottomrule
\end{tabular}
\vspace{-5pt}
\end{table}

The comparative analysis of proximity-based scores for the M1 and M2 models (Table \ref{tab:proximity_comparison}) reveals significant differences in their explanation alignment with the ground truth. M1 generally shows higher mean proximity scores and greater variability across all categories, indicating that its predictions are spatially farther from the true flare regions. In contrast, M2 demonstrates lower mean scores with reduced variability, reflecting more accurate and consistent predictions. Specifically, M2 outperforms M1 in false negatives, false positives, and true negatives, highlighting its improved alignment with actual flare regions. True positive scores for M2 further underscore its better spatial accuracy, making it more reliable for solar flare prediction. The box plots (Fig. \ref{fig:proximity_scores_boxplot}) illustrate the distribution of proximity scores across different categories, offering further insights. M1 displays higher median scores and greater variability, suggesting less precise predictions. Conversely, M2 exhibits lower medians and fewer extreme outliers, indicating closer and more consistent predictions. For true negatives, M2 achieves significantly lower median proximity scores, showcasing superior alignment with actual flare regions. Both models perform similarly for false positives and false negatives, but M2’s reduced variability and fewer outliers underscore its consistency.

\subsection{Comparative Analysis of ACR Across Models}
\vspace{-10pt}
\begin{table}[htbp]
\centering
\caption{Summary of Attribution Colocation Ratios for M1 and M2}
\label{tab:colocation_comparison}
\begin{tabular}{lcccc}
\toprule
\textbf{Metric} & \multicolumn{2}{c}{\textbf{Mean}} & \multicolumn{2}{c}{\textbf{Standard Deviation}} \\ 
 & \textbf{M1} & \textbf{M2} & \textbf{M1} & \textbf{M2} \\ \midrule
\textbf{FN} & 0.514 & 0.521 & 0.311 & 0.310 \\ 
\textbf{FP} & 0.511 & 0.446 & 0.289 & 0.273 \\ 
\textbf{TN} & 0.561 & 0.562 & 0.305 & 0.328 \\ 
\textbf{TP} & 0.574 & 0.463 & 0.306 & 0.342 \\ 
\bottomrule
\end{tabular}
% \vspace{-10pt}
\end{table}
The ACR analysis for the M1 and M2 models (Table \ref{tab:colocation_comparison}) highlights notable differences in prediction accuracy and consistency. M1 generally exhibits higher mean ACRs across most categories, suggesting that its predictions are more closely aligned with actual flare regions. However, M2 demonstrates less variability in these ratios, indicating a more consistent alignment across its predictions. The comparative mean and standard deviation values provide further insight into each model's relative performance and consistency in predicting solar flare locations. The box plots (Fig. \ref{fig:colocation_ratio_boxplot}) offer additional detail on the distribution of ACRs. While both models display similar medians across all categories, M2 shows narrower interquartile ranges (IQRs) for false positives and true negatives, reflecting more consistent alignment. In contrast, M1 displays greater variability in true positives and false negatives, suggesting less consistent predictions in these categories. For true positives, M2 achieves a lower median ACR but with reduced variability, implying consistent but slightly less accurate predictions compared to M1. For false negatives, M1 shows a slightly higher median but greater variability, further underscoring its inconsistency. Overall, M1 has higher median ACRs in some categories but less consistency due to greater variability, while M2 is more consistent with reduced variability.

\section{Conclusion} \label{sec:conclusion}
This study introduces a novel and fully automated method for analyzing explanations from full-disk models in a near-real-time setting. This approach represents a significant step forward in seamlessly integrating explainability into operational systems, enhancing the transparency and reliability of deep learning models for solar flare prediction. We utilized this method to evaluate the explanations generated from two CNN-based models and compared the reliability of these models. The findings underscore the potential of these explainable methods to bridge the gap between advanced model outputs and actionable insights. Future work should prioritize refining post-hoc explanation techniques, with an emphasis on mitigating noise during preprocessing, to further improve model interpretability and operational utility.\\

\noindent \textbf{Acknowledgements:}This work is supported by the National Science Foundation under Grant \#2104004. The data used in this study is a courtesy of NASA/SDO and the AIA, EVE, and HMI
science teams, and the NOAA (NGDC).

\bibliographystyle{IEEEtran} 
\bibliography{references}

\end{document}